\documentclass[conference]{IEEEtran}
\IEEEoverridecommandlockouts

\usepackage{cite}
\usepackage{amsmath,amssymb,amsfonts}
\usepackage{algorithmic}
\usepackage{graphicx}
\usepackage{subcaption}
\usepackage{float}
\usepackage{textcomp}
\usepackage{xcolor}
\def\BibTeX{{\rm B\kern-.05em{\sc i\kern-.025em b}\kern-.08em
    T\kern-.1667em\lower.7ex\hbox{E}\kern-.125emX}}
\begin{document}

\title{Joint Optimization of Fleet Size, Localization
Infrastructure, and Trajectories for Multi-AUV
Persistent Monitoring\\
}

\author{
\IEEEauthorblockN{
Jan Mikołajczyk\IEEEauthorrefmark{1}\IEEEauthorrefmark{2},
Jakub Bałenkowski\IEEEauthorrefmark{2},
Maciej Borowiecki\IEEEauthorrefmark{1},
Jan Ciecierski\IEEEauthorrefmark{1},\\
Marcel Kolecki\IEEEauthorrefmark{2},
Julian Maleta\IEEEauthorrefmark{2},
Maksymilian Bernacki\IEEEauthorrefmark{2}
}

\IEEEauthorblockA{\IEEEauthorrefmark{1}
University of Warsaw, Faculty of Physics / Faculty of Mathematics, Informatics and Mechanics, Warsaw, Poland
}

\IEEEauthorblockA{\IEEEauthorrefmark{2}
Warsaw University of Technology, Faculty of Power and Aeronautical Engineering\\ / Faculty of Electronics and Information Technology, Warsaw, Poland
}
\thanks{This work was supported by the Polish Ministry of Science and Higher Education under the ``Student Research Groups Create Innovations'' (SKNTI) program, project no. SKN/SP/658593/2026.}
\thanks{Corresponding author: jt.mikolajc2@student.uw.edu.pl}
\thanks{Published in OCEANS 2026 Sanya, IEEE.
DOI: 10.1109/OCEANS66983.2026.11617047.
\copyright~2026 IEEE. Personal use of this material is permitted.
Permission from IEEE must be obtained for all other uses, in any current or
future media, including reprinting/republishing this material for advertising
or promotional purposes, creating new collective works, for resale or
redistribution to servers or lists, or reuse of any copyrighted component of
this work in other works.}
}

\maketitle

\begin{abstract}
Persistent monitoring of large-scale subsea infrastructure, such as offshore wind farm foundations, export cables, and inspection corridors, remains a fundamental challenge for autonomous underwater systems.
The problem is
not solely spatial coverage, but rather temporal coverage: ensuring that all relevant regions of interest are revisited
within bounded time intervals under limited sensing, navigation uncertainty, and constrained vehicle resources.
We compare three specific classes of trajectory strategies:
\begin{enumerate}
\item Deterministic lane-based coverage (lawnmower patterns),
\item Stochastic exploration (biased random walk with boundary constraints),
\item Coordinated partition-based strategies (Voronoi tessellation)
\end{enumerate}
Rather than seeking a universally superior method, our objective is to evaluate how varying levels of localization accuracy impact temporal coverage across two distinct operational profiles: persistent monitoring of the entire operational workspace versus targeted inspection of critical infrastructure.
This paper presents a simulation-based, quantitatively grounded study of temporal coverage performance in multi-AUV persistent monitoring, explicitly disentangling and hierarchically evaluating three interacting design
dimensions: (i) trajectory generation strategy, (ii) fleet size, and (iii) localization accuracy provided by fixed acoustic beacon infrastructure.

To evaluate this, we introduce cost-aware evaluation framework combining fleet size,
number beacons, and mission energy expenditure into a unified cost function.
Results demonstrate that beyond a certain accuracy threshold, improvements in localization yield diminishing returns compared to coordinated trajectory design, an effect often obscured in integrated navigation-planning approaches.
Specifically, experiments conducted in a simulated operational environment reveal that coordinated-partition trajectories enable a reduced fleet size to achieve superior temporal coverage at a 39-46\% lower overall life-cycle cost  compared to rigid deterministic paths unoptimized for the localization infrastructure.

\end{abstract}

\section{Introduction}
The rapid expansion of the offshore energy sector has increased the importance  of the cost-effective subsea infrastructure maintenance. For offshore wind farms,
Operations and Maintenance (O\&M) costs can account for up to 30\% of the total life cycle
expenditure \cite{rockmann2017operation}. Efficient monitoring of export cables, turbine foundations, and inspection
corridors is therefore critical to reducing these costs and preventing catastrophic failures.
While Autonomous Underwater Vehicles (AUVs) offer a scalable solution for these tasks, the fundamental challenge remains ensuring temporal coverage rather than mere spatial reach.
Effective persistent monitoring depends on regularly revisiting target areas within
specific time frames to guarantee the prompt identification of anomalies. Success in these
missions requires a careful synchronization of three core elements: AUV swarm coordination
and path-planning logic, the overall scale of the fleet, and the precision of the navigation
systems, which typically utilize dead-reckoning and acoustic beacons.
These challenges manifest differently depending on the operational objective: full workspace coverage versus targeted inspection of critical infrastructure such as turbine foundations.
Existing research frequently evaluates these parameters in isolation, often neglecting the
complex trade-offs that occur in dynamic underwater environments characterized by
stochastic disturbances and limited communication. This paper presents a quantitatively
grounded study that explicitly disentangles these dimensions. We test the hypothesis that
beyond a specific localization accuracy threshold, further hardware investment yields
diminishing returns compared to optimized trajectory design. Through the use of a lightweight 2D simulation framework equipped with spatial ocean current models, we provide actionable insights into the design of economically viable multi-AUV monitoring systems.
The primary contributions of this work are:
\begin{enumerate}
    \item A systematic evaluation of trajectory, fleet, and localization effects on the primary temporal metric (AoI), with IVT as a secondary diagnostic indicator.
    \item The identification of accuracy thresholds where localization improvements reach a point of saturation, depending on chosen path-planning strategy.
    \item A cost-aware framework to optimize the balance between sensing infrastructure and
fleet size.
\end{enumerate}

\section{Related Work}

\subsection{Persistent Monitoring and Temporal Coverage}
Persistent monitoring relies on strict temporal guarantees rather than single-pass spatial inspections. This problem is traditionally modeled using scalar fields \cite{smith2011persistent} or objective analysis cost functions \cite{leonard2007collective}. Recently, the Age of Information (AoI) metric has become central to evaluating temporal coverage, optimized via deep reinforcement learning \cite{alhabob2023age} or metaheuristics \cite{cao2023joint}. However, these approaches typically assume idealized environments, leaving the impact of accumulating navigation drift on AoI reliability largely underexplored.

\subsection{Multi-AUV Swarm Coordination}
Multi-AUV coordination must overcome severe acoustic limitations, for instance, through periodic connectivity \cite{hollinger2012multi}. Workspace partitioning frequently utilizes Voronoi diagrams \cite{zhou2023applications}, with advanced models addressing fleet heterogeneity \cite{guo2023static} and kinematic safety via Control Barrier Functions \cite{mao2023cooperative}. Despite these algorithmic advancements, there is a lack of systematic studies quantifying the extent to which coordinated strategies outperform classical patterns (e.g., lawnmower algorithms) under realistic conditions of severe noise and drift.

\subsection{Navigation and Localization Uncertainty}
Underwater navigation fundamentally relies on Dead Reckoning (DR) aided by sparse acoustic networks (e.g., LBL) and nonlinear estimation filters \cite{paull2014auv}. Because trajectory planning dictates physical operational capabilities, modern algorithms increasingly incorporate ocean current models to reduce mission time \cite{garau2005path} and explicitly model bathymetry to avoid acoustic shadows \cite{almuzaini2024trajectory}.

\subsection{The Gap: System-Level Integration}
While existing literature extensively covers isolated sub-problems, it lacks integrated studies evaluating the joint influence of trajectory strategy, fleet size, and localization accuracy. Our work fills this gap. By utilizing a simulation environment that incorporates stochastic ocean currents and acoustic localization errors, we explicitly investigate the saturation thresholds of the system. This approach reveals critical engineering trade-offs, highlighting the exact points where investments in navigation infrastructure yield diminishing returns for temporal coverage compared to coordinated motion planning.

\section{Problem Formulation and Evaluation Metrics}

\subsection{System Model and Assumptions}
Let the subsea environment be defined as a bounded workspace, discretized into a set of spatial cells $C = \{c_1, c_2, \dots, c_N\}$. A subset of these cells may be designated as critical Regions of Interest (RoIs) requiring persistent observation. The monitoring mission is executed by a homogeneous fleet of Autonomous Underwater Vehicles (AUVs), denoted as $\mathcal{A} = \{a_1, a_2, \dots, a_k\}$, operating over a mission horizon $T_{max}$. 

The vehicles operate under strict kinematic and environmental constraints. The true position of vehicle $a_k$ at time $t$ is subject to external disturbances, primarily modeled as low-to-moderate ocean currents $\vec{d}(x,y,t)$ with magnitudes on the order of $0.1$ to $0.3$~m/s. Furthermore, localization is not perfect; the estimated position relies on a fixed acoustic beacon infrastructure, incorporating zero-mean Gaussian noise with an measured with RMSE. 

\subsection{Formal Problem Statement}
The core objective of the multi-AUV persistent monitoring task is not merely single-pass spatial coverage, but maintaining strict temporal coverage guarantees under operational uncertainty. Formally, the problem requires generating a set of continuous trajectories $\mathcal{T} = \{\tau_1, \tau_2, \dots, \tau_k\}$ for the fleet $\mathcal{A}$ that minimizes the temporal gaps between consecutive visits to all cells $c_i \in C$, while simultaneously minimizing system-level resource expenditure (energy) and maintaining robustness against environmental disturbances. Depending on the operational profile, this objective applies either to the full workspace $C$ or exclusively to the designated RoI subset.

\subsection{Performance Metrics}
To quantitatively evaluate and disentangle the performance of deterministic, stochastic, and coordinated trajectory strategies, we define the following set of primary and secondary metrics:

\subsubsection{Temporal Coverage Metrics}
\begin{itemize}
    \item \textbf{Age of Information (AoI):} The fundamental metric for temporal coverage. For any given cell $c_i$ at time $t$, $AoI(c_i, t)$ is defined as the elapsed time since its most recent visit by any vehicle in the fleet. If $t_{last}(c_i)$ denotes the timestamp of the last visit, it is expressed as:
    \begin{equation}
        AoI(c_i, t) = t - t_{last}(c_i)
    \end{equation}
    The mean AoI across the grid directly dictates the temporal bounds of the monitoring system.
    
    \item \textbf{Inter-Visit Time (IVT):} While AoI captures the instantaneous state, IVT measures the duration between two consecutive visits to the same cell $c_i$. 
    IVT must be interpreted alongside spatial coverage metrics, as high revisit frequency concentrated in a small area can produce deceptively low IVT values without meaningful monitoring progress.
    
\end{itemize}

\subsubsection{Resource Utilization}
\begin{itemize}
    \item \textbf{Energy Proxy ($Ep$):} To capture the resource footprint of each strategy, 
    we formulate an energy proxy based on the relative kinematic effort. 
    The energy footprint is defined as the time integral of the cubic relative linear velocity $v{rel}(t)$ and squared angular velocity $\omega(t)$ for all $K$ AUVs:
    \begin{equation}
        \begin{split}
        E_p(\mathcal{T}) &= \sum_{k=1}^{K} \int_{0}^{T_{max}} \Big( w_v |v_{rel, k}(t)|^3 \\
        &\quad + w_\omega |\omega_k(t)|^2 \Big) dt
        \end{split}
    \end{equation}
    where $v{rel, k}(t) = v_k(t) - \vec{d}(x,y,t)$ is the AUV's velocity relative to the local ocean current $\vec{d}(x,y,t)$, and $w_v$, $w_\omega$ are hydrodynamic weighting factors for forward propulsion and rotational maneuvers. This model explicitly penalizes strategies that force AUVs to navigate directly against strong ocean currents or perform frequent, sharp course corrections compared to smoother Voronoi relaxations. The empirical values of these weights are modeled after standard torpedo-shaped autonomous underwater vehicles (e.g., the REMUS 600 class \cite{prestero2001verification}), reflecting the realistic physical constraints of fluid resistance and turning effort.
\end{itemize}

\subsection{The Primacy of Temporal Metrics}
While spatial metrics describe observed areas, temporal metrics dictate data freshness and reliability, making them critical for persistent monitoring. Consequently, this study prioritizes Age of Information (AoI) as the primary performance indicator. AoI explicitly quantifies temporal gaps, ensuring that bounding its maximum value across Regions of Interest (RoIs) guarantees no critical infrastructure remains uninspected beyond safe operational limits. Inter-Visit Time (IVT) serves only as a supplementary diagnostic metric.

\subsection{Operational Environment and Wind Farm Topology}
To maximize the practical applicability of our study, the simulation environment is modeled after a representative offshore wind farm topology. Consequently, the operational workspace is defined as an asymmetric trapezoidal polygon (with maximum spans of approximately $7.9$~km by $7.7$~km) encompassing $13$ turbine foundations explicitly designated as our Regions of Interest (RoIs).

\subsection{System Constants and Optimal Baselines}
To isolate the steady-state performance of the evaluated trajectory algorithms, several environmental and mission parameters are fixed to their optimal baselines:
\begin{itemize}
    \item \textbf{Initial State \& Infrastructure:} AUV starting positions are pre-computed, and a fixed, heuristically optimized acoustic beacon layout is assumed.
    \item \textbf{Kinematics \& Environment:} AUV surge speed is held constant at $1.3$~m/s with ambient ocean currents ranging from $0.1$~m/s to $0.3$~m/s.
\end{itemize}

\subsection{Control Variables (Parameters)}
Our experimental matrix explicitly evaluates combinations of the following parameters:
\begin{itemize}
    \item \textbf{Trajectory Strategy ($\mathcal{T}$):} Deterministic (Lawnmower), Stochastic (Biased Random Walk), and Coordinated (Voronoi tessellation).
    \item \textbf{Fleet Size ($K$):} Scaled incrementally from $6$ to $10$ homogeneous vehicles.
    \item \textbf{RoI Prioritization Schema:} Uniform workspace coverage vs. targeted inspection of specific turbine foundations.
\end{itemize}

\section{Trajectory Generation Strategy}

\subsection{Notation and Information state assumption}
Let $\mathbf{p}_k(t) = [x_k(t), y_k(t)]$ denote the current position vector of the $k$-th AUV time $t$ and let $\delta$ represent a constant scalar distance covered in a single decision step.

To isolate the influence of trajectory geometry on temporal coverage and cost optimization, this study assumes periodic, error-free state synchronization among the fleet within a limited communication range, isolating the effect of trajectory geometry from communication uncertainty. At any decision time $t$, the $k$-th AUV possesses a local information state $\mathcal{X}_k$, defined as:
\begin{equation}
\mathcal{X}_k = \big\{ \mathbf{H}_k, \ \mathbf{P}_{neighbors}, \ \mathcal{B}, \ \mathcal{V}_{\text{visits}} \big\}
\end{equation}
where $\mathbf{H}_k$ represents the discrete history of the vehicle's own past positions, $\mathbf{P}_{neighbors}$ contains the current estimated positions of other AUVs within the acoustic modem range, $\mathcal{B}$ is the map of all acoustic beacons, and $\mathcal{V}_{\text{visits}}$ is a memory structure logging the exact timestamp of recent inspections at target infrastructure (e.g., wind turbines).
To enable decentralized coordination, the AUV fleet relies on Medium Frequency (MF) acoustic modems. While theoretical limits may be higher, the shallow-water environment and active offshore infrastructure introduce severe multipath propagation and ambient acoustic noise. The reliable communication radius is bounded to $R_{\text{comm}} = 1500$ m, aligning with the empirical performance of industry-standard equipment \cite{freitag2005}.

\subsection{Stochastic exploration (biased random walk)}
To determine its next waypoint, the AUV chooses from $M$ uniformly distributed candidate positions located at a distance $\delta$. The discrete set of available heading angles is defined as:
\begin{equation}
\Theta = \Bigl\{ \theta_j\ | \text{ } \theta_j = j \frac{2\pi}{M}, \text{ } j \in \{1, \dots,M\} \Bigr\} 
\end{equation}
For each potential direction $\theta_j$, the predicted AUV position is:
\begin{equation}
\hat{\mathbf{p}}_{k,j} = \mathbf{p}_k(t) + \begin{bmatrix} \delta \cos\theta_j \\ \delta \sin\theta_j \end{bmatrix}.
\end{equation}
For every  $\hat{\mathbf{p}}_{k,j}$  its viability is calculated with score function. Score function evaluates each $\hat{\mathbf{p}}_{k,j}$ value as a linear combination of $M$ independent bias criteria:
\begin{equation}
S(\theta_j) = \sum_{m=1}^{M} w_m \cdot B_m(\hat{\mathbf{p}}_{k,j}, \mathcal{X}_k),
\end{equation}
where $w_m$ is weight of $m$-th bias. 
\begin{equation}
P(\theta_j) = \frac{\exp \left( \frac{S(\theta_j)}{\tau} \right)}{\sum_{m=1}^{M} \exp \left( \frac{S(\theta_m)}{\tau} \right)},
\end{equation}
 Probability of $\theta_j$ direction is modeled using Softmax function, where  $\tau > 0$ is a temperature parameter balancing stochastic diversity and bias-driven exploitation. 

\subsection{Deterministic lane-based coverage (lawnmower patterns)}
As a standard for subsea inspections, the deterministic lawnmower strategy is evaluated. In this approach, the operational space is partitioned into $N$ equal sub-regions assigning a disjoint operational zone to each AUV.
The trajectory for the $k$-th AUV is generated offline as a sequence of waypoints instead of dynamically looking for next move or direction. It is important to distinguish between trajectory tracking and trajectory generation. While the vehicle corrects its physical location using acoustic beacons, the sequence of target waypoints is entirely decoupled from the vehicle's real-time information state. The AUV only alters its global path to maintain selected waypoints, but it does not respond to growing localization uncertainty. In acoustically sparse regions, the AUV's inability to autonomously seek triangulation causes navigational drift and compromises temporal coverage.

\subsection{Partition based strategies (biased Voronoi tessellation)}
The advantage of partition-based strategies is the spatial decoupling. By assigning each AUV its own dynamic region of responsibility (Voronoi cell), the algorithm enforces the division of labor and prevents more than one AUV to explore the same region in the same moment.
The Voronoi cell $\Omega_k$ assigned to the $k$-th AUV is defined as the set of all points $\mathbf{q} =(x,y)$ that are strictly closer to $\mathbf{p}_k(t)$ than to any other AUV's position $\mathbf{p}_j(t)$ currently visible in communication range:
\begin{equation}
\begin{split}
\Omega_k = \{ \mathbf{q} \mid ||\mathbf{q} - \mathbf{p}_k(t)|| < ||\mathbf{q} - \mathbf{p}_j(t)||, \\
\forall k \neq j , |\mathbf{p}_k(t)-\mathbf{p}_j(t)|<R_{comm}\}
\end{split}
\end{equation}
Standard CVT algorithms eventually cause AUVs to stagnate at their cell centroids, halting continuous coverage.
To overcome this stagnation, the proposed algorithm abandons moving towards center of mass. Instead, it leverages the same point evaluation function $S(\mathbf{q})$ introduced in the stochastic strategy. In this deterministic approach, the $k$-th AUV performs a discrete
grid search within a local search horizon in its cell $\Omega_k$ to identify the target point $\mathbf{q}_{k}^{*}$ that yields the maximum score, and satisfies the condition:
\begin{equation}
S(\mathbf{q}^*_k, \mathcal{X}_k) = \max_{\mathbf{q} \in \Omega_k} S(\mathbf{q}, \mathcal{X}_k)
\end{equation}
Once the most desirable point $\mathbf{q}^*_k$ is identified, the AUV performs a step toward it. To provide a mechanism for escaping local utility extrema, the sequential waypoint $\hat{\mathbf{p}}_k$ is determined via a piecewise function:
\begin{equation}
\hat{\mathbf{p}}_k = \begin{cases} \mathbf{p}_k(t) + \delta \frac{\mathbf{q}^*_k - \mathbf{p}_k(t)}{||\mathbf{q}^*_k - \mathbf{p}_k(t)||} & \text{if } ||\mathbf{q}^*_k - \mathbf{p}_k(t)|| > \epsilon \\ \mathbf{p}_k(t) + \delta_{min} \mathbf{u}_{rand} & \text{otherwise} \end{cases}
\end{equation}
where $\epsilon$ is the minimal distance threshold (e.g., $0.1$ m) below which the AUV is considered to have reached the target point, $\mathbf{u}_{rand}$ is a uniformly sampled random unit vector, and $\delta_{min}$ represents a fixed minimum step size. This conditional logic ensures the AUV maintains momentum and keeps exploring even when centered directly over a local utility peak.

\subsection{Kinematic Constraints and Collision Avoidance}
To ensure kinematic realism and accurate energy modeling, AUVs physically decelerate during sharp maneuvers. The actual step size $\delta_{actual}$ is dynamically scaled based on the required heading change $\Delta\theta$:
\begin{equation}
\delta_{actual} = \delta_{max} \left( \eta_{min} + (1 - \eta_{min}) \frac{1 + \cos(\Delta\theta)}{2} \right)
\end{equation}
where $\eta_{min}$ defines the minimum velocity factor. Furthermore, a deterministic low-level obstacle avoidance layer prevents collisions and deadlocks. If safety margins are violated, the AUV sequentially halves its speed, applies a $45^\circ$ right-hand steering rule, or executes stochastic micro-displacements to find a safe vector.

\subsection{Biases and Implementation Constraints}

\subsubsection{Kinematic Momentum}
To prevent twisting movements, the momentum bias rewards candidate points that align with the AUV's current velocity vector. It is calculated as the cosine similarity between the previous movement step $\mathbf{v}_{prev}$ and the proposed direction $\mathbf{q} - \mathbf{p}_k(t)$:
\begin{equation}
B_{mom}(\mathbf{q}) = \frac{\mathbf{v}_{prev} \cdot (\mathbf{q} - \mathbf{p}_k(t))}{||\mathbf{v}_{prev}|| \ ||\mathbf{q} - \mathbf{p}_k(t)||}
\end{equation}

\subsubsection{Swarm Dispersion}
Applied exclusively in the Random Walk strategy (due to the lack of Voronoi boundaries), this bias promotes spatial distribution. It computes a cumulative escape vector $\vec{e}_{disp}$ away from all neighbors within the communication range and rewards candidate directions aligned with this escape vector via cosine similarity:
\begin{equation}
\vec{e}_{disp} = \sum_{j \in \mathcal{N}_k} (\mathbf{p}_k(t) - \mathbf{p}_j(t)), \quad B_{avoid}(\mathbf{q}) = \frac{\vec{e}_{disp} \cdot (\mathbf{q} - \mathbf{p}_k(t))}{||\vec{e}_{disp}|| \ ||\mathbf{q} - \mathbf{p}_k(t)||}
\end{equation}

,where $\mathcal{N}_k$ is a list of neighbors within $R_{comm}$.

\subsubsection{Transponder Localization \& Triangulation}
To maintain the AUV within acoustically observable areas, the algorithm utilizes directional pull vectors. The patrol bias pulls the vehicle toward the nearest transponder (ignoring the most recently visited one to enforce continuous movement). Let $\mathbf{P}_{patrol} = \mathbf{p}_{b\_nearest} - \mathbf{p}_k(t)$ be the pull vector toward the closest valid beacon:
\begin{equation}
B_{patrol}(\mathbf{q}) = \frac{\mathbf{P}_{patrol} \cdot (\mathbf{q} - \mathbf{p}_k(t))}{||\mathbf{P}_{patrol}|| \ ||\mathbf{q} - \mathbf{p}_k(t)||}
\end{equation}
Similarly, if the AUV detects fewer than 3 beacons, a triangulation bias calculates the spatial centroid $\mathbf{p}_{tri}$ of all currently visible beacons. It generates a pull vector $\mathbf{P}_{tri} = \mathbf{p}_{tri} - \mathbf{p}_k(t)$ to promote entry into high-accuracy zones:
\begin{equation}
B_{triang}(\mathbf{q}) = \frac{\mathbf{P}_{tri} \cdot (\mathbf{q} - \mathbf{p}_k(t))}{||\mathbf{P}_{tri}|| \ ||\mathbf{q} - \mathbf{p}_k(t)||}
\end{equation}

\subsubsection{Cyclic Target Inspection}
For scenarios demanding persistent monitoring of specific targets (e.g., wind turbines), a temporal cycle mechanism is introduced. Instead of permanent attraction, a target $\mathcal{T}_k$ is considered active only if the time elapsed since the last visit exceeds a defined $t_{cooldown}$. Let $\mathbf{P}_{turb} = \mathcal{T}_{active\_nearest} - \mathbf{p}_k(t)$ be the pull vector toward the closest active target:
\begin{equation}
B_{turb}(\mathbf{q}) = \frac{\mathbf{P}_{turb} \cdot (\mathbf{q} - \mathbf{p}_k(t))}{||\mathbf{P}_{turb}|| \ ||\mathbf{q} - \mathbf{p}_k(t)||}
\end{equation}

\subsubsection{Memory Penalty}
Discourages the AUV from revisiting its recently traversed path to maintain forward exploration momentum. It applies an exponentially decaying penalty based on the vehicle's recorded position history $\mathbf{H}_i$, capped at a normalized minimum of $-1.0$:
\begin{equation}
    B_{hist}(\mathbf{q}) = \max\left(-1, -\sum_{h \in \mathbf{H}_i} \exp\left(-\frac{||\mathbf{q} - h||}{\lambda_{\text{falloff}}}\right) \right)
\end{equation}

\subsection{Hyperparameter Tuning}
The overall performance of the trajectory strategies relies on the weights ($w_m$) assigned to each bias function, as well as internal parameters (e.g., $T_{\text{cooldown}}$, $\lambda_{\text{falloff}}$). These parameters are tuned to be as effective as possible offline using the Tree-structured Parzen Estimator (TPE) algorithm via the Optuna framework. 

For a single simulation run, the objective function strictly penalizes unachieved spatial coverage while minimizing the Age of Information (AoI):
\begin{equation}
\mathcal{L}(\mathbf{w}) = \gamma_{aoi} \cdot \overline{AoI} + \gamma_{cov} \cdot \max(0, 100 - C_{actual})^2
\end{equation}
where $\overline{AoI}$ is the mean Age of Information across the grid, $C_{\text{actual}} \in [0,100]$ is the achieved coverage percentage, and $\gamma_{aoi}, \gamma_{cov}$ are scaling multipliers.

To prevent the optimizer from overfitting the weights $w_m$ to a single environmental disturbance, the objective function is evaluated stochastically. For every hyperparameter trial, the simulation is executed across a set of independent random seeds $\mathcal{S}$, each generating unique ocean current distributions. The final loss is the expected value across all seeded environments:
\begin{equation}
\mathcal{J}_{\text{Optuna}}(\mathbf{w}) = 
\frac{1}{|\mathcal{S}|} \sum_{s \in \mathcal{S}} 
\mathcal{L}_s(\mathbf{w}).
\end{equation}

\subsection{Odometry Tracking with Partial Acoustic Assimilation}
To address sparse acoustic updates efficiently, we replace complex full-state EKFs with a heuristically scaled assimilation filter. The AUV's internal dead-reckoning fuses a compass ($3^\circ$ noise) and DVL ($2\%$ velocity error, $5\%$ current error) \cite{kinsey2006survey, paull2014auv}. The state covariance is updated based on the number of visible beacons ($n$):

\textbf{Case 1 ($n \ge 3$):} Full triangulation overrides the internal estimate and resets variance to a minimal baseline.

\textbf{Case 2 ($n = 2$):} The state is corrected using a distance-dependent measurement variance $R = (\sigma_0 + \beta \cdot d_{avg})^2$. Based on typical LBL accuracy \cite{kinsey2006survey, paull2014auv}, the baseline resolution is $\sigma_0 = 0.5$~m, while $\beta = 0.005$ accounts for acoustic attenuation over long ranges \cite{webster2012}.

\textbf{Case 3 ($n = 1$):} A synthetic measured point is created by projecting scalar distance along the compass vector. To maintain filter consistency under weak observability, severe covariance inflation ($R_{inflated} \approx 64 \cdot R_{base}$) is strictly applied before the Kalman update \cite{webster2012}.

\section{Optimal Acoustic Infrastructure Placement}
The spatial arrangement of acoustic transponders dictates the localization accuracy of the fleet, which directly impacts the drift and temporal coverage metrics. Therefore, the placement of $K$ beacons is formulated as an optimization problem bounded by the infrastructural map limits.

\subsubsection{Objective Function and CRLB}
Unlike idealized models that strictly require triangulation from three sources, state-of-the-art navigation filters can assimilate partial updates. Thus, our layout optimizer deliberately rewards partial spatial coverage to force the expansion of the acoustic network towards the edges of the operational space. 

Let $\rho_1, \rho_2,$ and $ \rho_3$ denote the ratio of the discrete map grid covered by at least one, two, and three beacons, respectively. The optimal layout is determined by minimizing a composite cost function $J_{total}$:
\begin{equation}
J_{total} = J_{geom} + J_{crlb} - J_{cov} + J_{bias}
\end{equation}

Due to the complex objective landscape, a Differential Evolution (DE) metaheuristic is employed to minimize $J_{total}$, defined by the following components:

\begin{itemize}
    \item \textbf{Coverage Reward ($J_{cov}$)}: $J_{cov} = \alpha_1 \rho_1 + \alpha_2 \rho_2 + \alpha_3 \rho_3$. Heavily weighting $\rho_2$ maximizes usable acoustic area without unnecessary central clustering.
    \item \textbf{Geometric Dilution \& CRLB ($J_{crlb}$)}: Sums the mean and 90th percentile of the Cramér-Rao Lower Bound trace (evaluated for $\ge 2$ beacons) to severely penalize collinear arrangements.
    \item \textbf{Hard Geometric Constraints ($J_{geom}$)}: Applies linear penalties for inter-beacon distances below $0.8 \cdot R_{beacon}$ to prevent redundant clustering while ensuring sufficient acoustic overlap \cite{ferreira2023sensor}. It also restricts 3-beacon intersection angles to $[45^\circ, 135^\circ]$ for robust triangulation.
    \item \textbf{Center Bias ($J_{bias}$)}: A soft regularization term pulling the acoustic grid toward the map center.
\end{itemize}
\begin{figure*}
    
    \centering

    \begin{subfigure}{0.32\textwidth}
        \includegraphics[width=\textwidth]{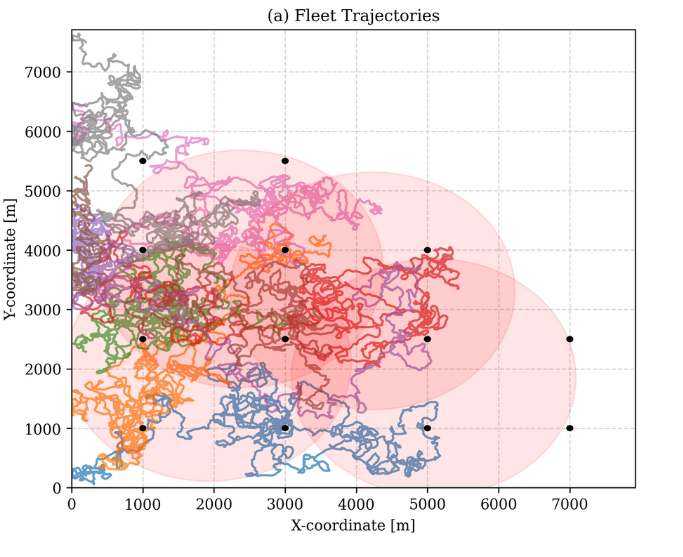}
        \caption{Biased random walk, global profile. }
    \end{subfigure}
    \hfill
    \begin{subfigure}{0.32\textwidth}
        \includegraphics[width=\textwidth]{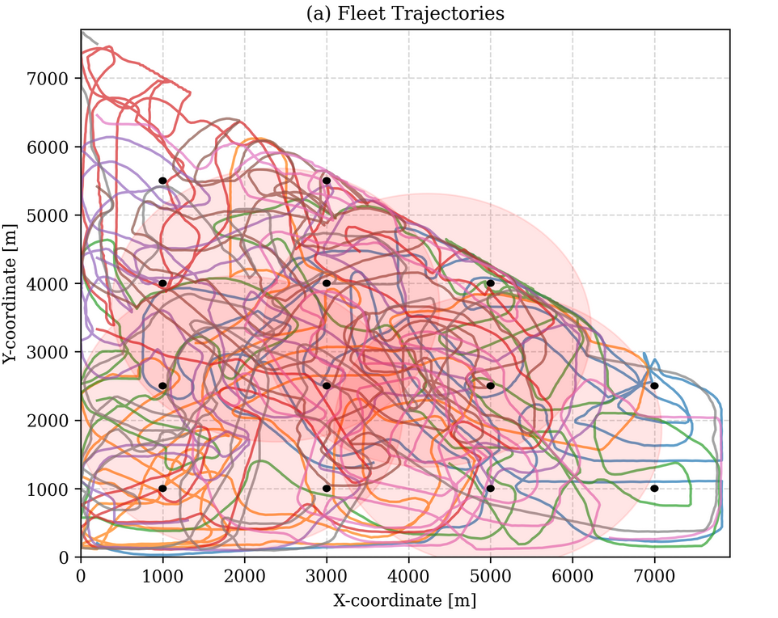}
        \caption{Voronoi-based, global profile.}
    \end{subfigure}
    \hfill
    \begin{subfigure}{0.32\textwidth}
        \includegraphics[width=\textwidth]{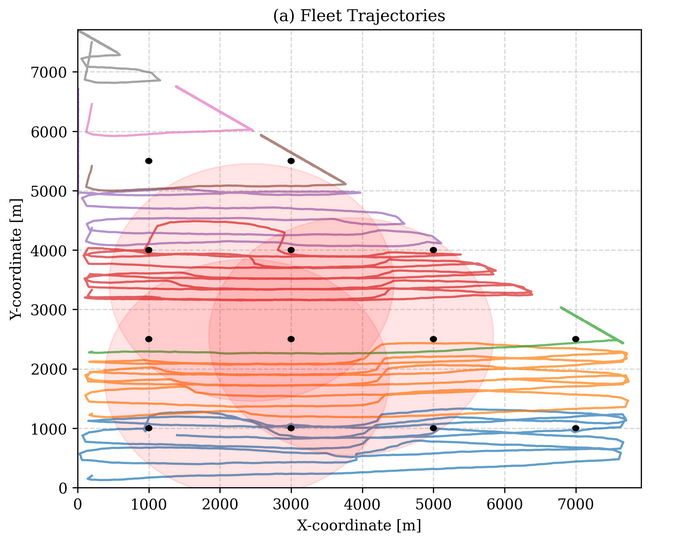}
        \caption{Lawnmower, global profile. }
    \end{subfigure}

    \vspace{0.5em}

    \begin{subfigure}{0.32\textwidth}
        \includegraphics[width=\textwidth]{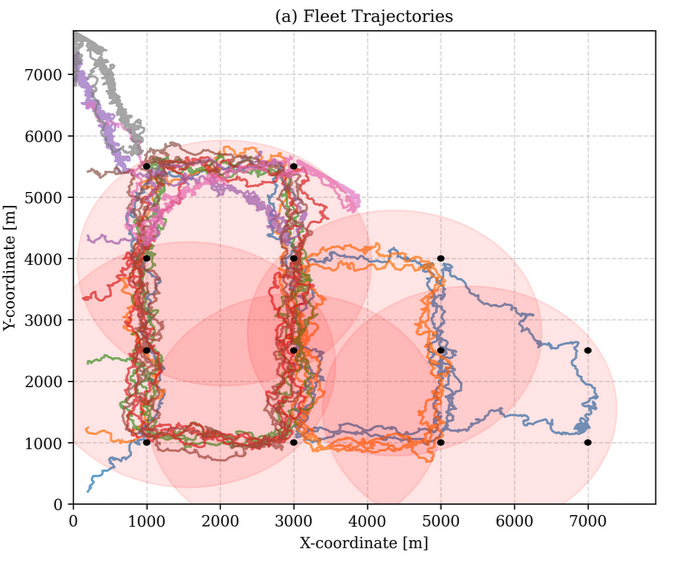}
        \caption{Biased random walk, turbine inspection 
profile.}
    \end{subfigure}
    \hfill
    \begin{subfigure}{0.32\textwidth}
        \includegraphics[width=\textwidth]{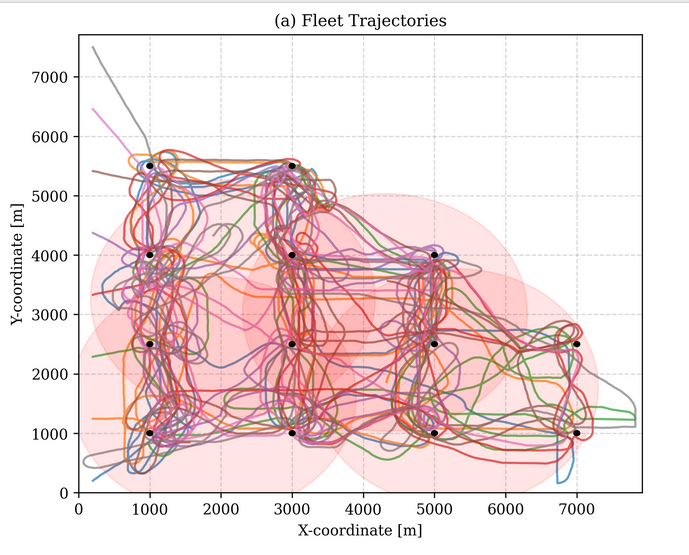}
        \caption{Voronoi-based, turbine inspection 
profile.}
    \end{subfigure}
    \hfill
    \begin{subfigure}{0.32\textwidth}
        \includegraphics[width=\textwidth]{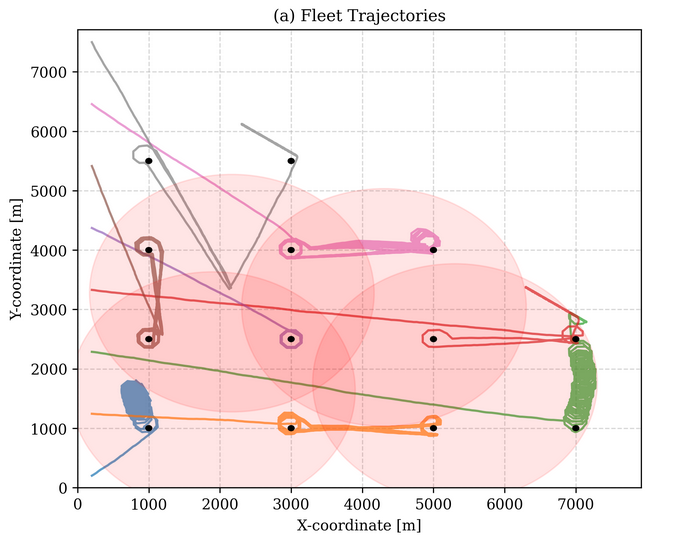}
        \caption{Lawnmower, turbine inspection profile.}
    \end{subfigure}

    \caption{Representative fleet trajectories ($N=8$ 
AUVs) for all three evaluated strategies under 
both operational profiles. Top row: global workspace 
coverage profile. Bottom row: turbine inspection 
profile. Columns correspond to biased random walk 
(left), Voronoi-based (center), and lawnmower 
(right) strategies. Black dots indicate turbine 
foundation locations. Pink shaded circles indicate 
acoustic beacon coverage ranges.}
    \label{fig:trajectories}
\end{figure*}

\subsubsection{Transponder Range}
Infrastructure is modeled using Medium Frequency (MF) LBL transponders. Reflecting their typical robust performance bounds \cite{kinsey2006survey}, the maximum triangulation range is strictly limited to $R_{beacon} = 2000$~m for a realistic deployment strategy.

\section{Methodology of Experiments}
\begin{figure*}[t]
\centering
\begin{subfigure}{0.48\textwidth}
    \includegraphics[width=\textwidth]
    {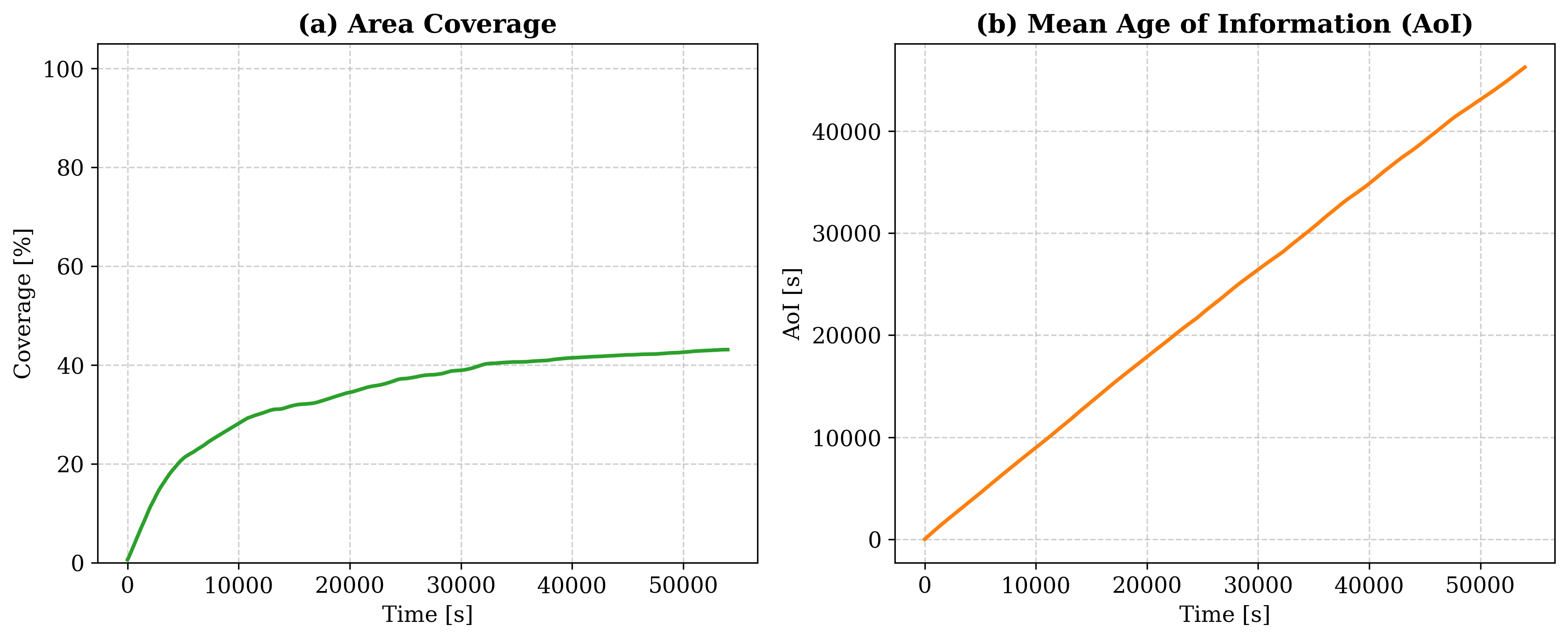}
    \subcaption{Lawnmower, $K=1$: coverage 
    plateaus at 43\,\%, AoI grows linearly.}
    \label{fig:lm_low}
\end{subfigure}
\hfill
\begin{subfigure}{0.48\textwidth}
    \includegraphics[width=\textwidth]
    {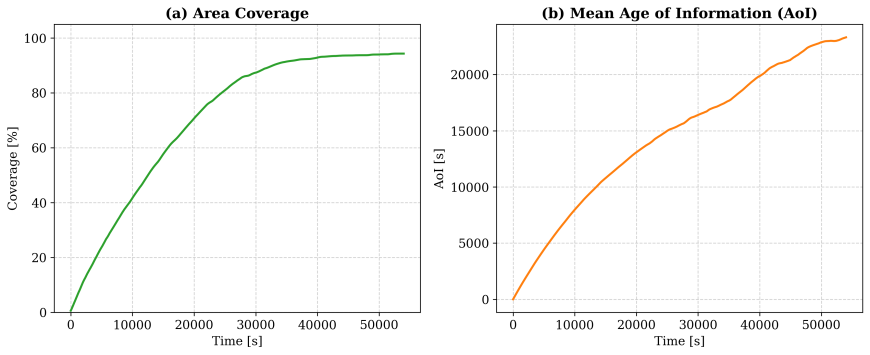}
    \subcaption{Lawnmower, $K=6$: coverage 
    reaches 94\,\%, AoI growth substantially 
    reduced.}
    \label{fig:lm_high}
\end{subfigure}

\vspace{0.3cm}

\begin{subfigure}{0.48\textwidth}
    \includegraphics[width=\textwidth]
    {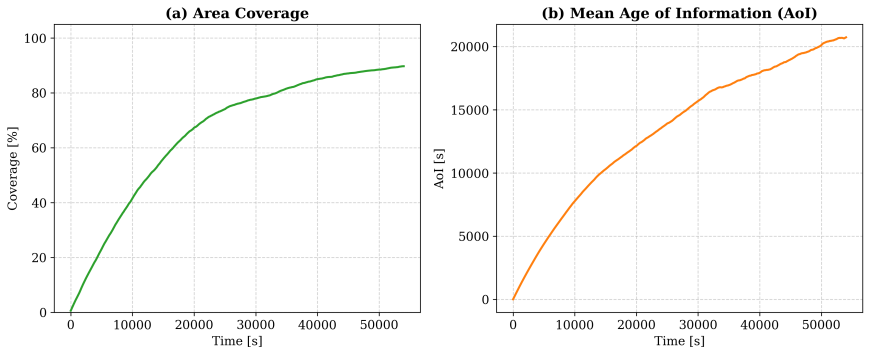}
    \subcaption{Voronoi-based, $K=1$: coverage 
    reaches 90\,\%, AoI comparable to $K=6$ 
    configuration.}
    \label{fig:vor_low}
\end{subfigure}
\hfill
\begin{subfigure}{0.48\textwidth}
    \includegraphics[width=\textwidth]
    {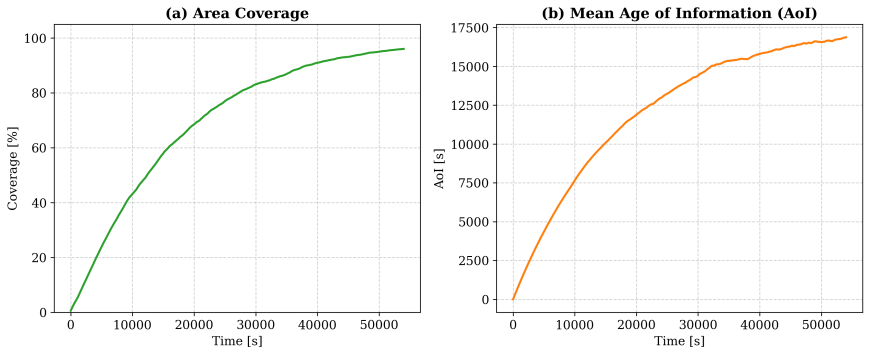}
    \subcaption{Voronoi-based, $K=6$: coverage 
    reaches 96\,\%, AoI marginally lower than 
    at $K=1$.}
    \label{fig:vor_high}
\end{subfigure}

\caption{Area coverage and mean AoI for lawnmower 
(top) and Voronoi-based (bottom) strategies under 
sparse ($K=1$) and dense ($K=6$) infrastructure 
($N=8$ AUVs, global profile).}
\label{fig:beacon_effect}
\end{figure*}

\subsection{Economic and Efficiency Evaluation}
To evaluate the economic trade-offs of different configurations, we define the life-cycle cost function ($\Phi_{total}$) and the Monitoring Efficiency Index ($\Psi$):

\begin{equation}
    \Phi_{total} = (K \cdot 15000 + B \cdot 18000) + N_{missions} \cdot (E_p \cdot c_e)
    \label{eq:unified_cost}
\end{equation}

where $K$ and $B$ denote the number of AUVs and acoustic beacons, with unit costs of \$15,000 and \$18,000, respectively. $N_{missions}$ is the estimated lifecycle mission count, $E_p$ is the Energy Proxy, and $c_e = 0.001$ scales the kinematic effort to subsea energy lifecycle costs.

To quantify the performance return on financial investment, we introduce the Monitoring Efficiency Index ($\Psi$):

\begin{equation}
    \Psi = \frac{\Phi_{total} \cdot \overline{AoI}}{10 \cdot C_{actual}}
    \label{eq:mei_index}
\end{equation}

where $\overline{AoI}$ is the mean Age of Information and $C_{actual}$ is the achieved coverage percentage. Minimizing $\Psi$ identifies configurations that yield the highest monitoring quality per dollar spent. (Divided by $10$ for clearer results).

\subsection{Simulation Environment and Hydrodynamic Modeling}
To maintain computational tractability while preserving the topological complexity of the reference project of offshore wind farm, physical dimensions were downscaled by a factor of $0.1$. 

Ocean currents are dynamically modeled using the 2D Shallow Water Equations over discrete steps ($\Delta t = 30.0$~s). The numerical scheme explicitly incorporates Coriolis forces, linear bottom friction, and stochastic surface wind perturbations. 

Operating within this environment, the AUV sensory footprint is modeled as a deterministic disc. The effective radius is set to $100$~m for global workspace exploration, and deliberately constrained to $50$~m for targeted infrastructure inspection to reflect the higher resolution required for structural anomaly detection \cite{wynn2014autonomous}.

\section{Results and Discussion}

\subsection{Trajectory Characterization}

Fig.~\ref{fig:trajectories} contrasts representative fleet trajectories ($N=8$). While baseline behaviors are evident under the global workspace profile—with Voronoi maintaining strict spatial separation and lawnmower executing systematic parallel lanes—the turbine inspection profile reveals pronounced behavioral divergence. Adaptive strategies (Voronoi and random walk) successfully cluster around the targeted foundations, whereas the deterministic lawnmower strategy suffers from severe navigational degradation, deviating significantly from its planned inspection paths.
\subsection{Impact of Localization Infrastructure and Saturation Thresholds}
Figs.~\ref{fig:beacon_effect} illustrate the 
contrasting sensitivity of the lawnmower and 
Voronoi-based strategies to acoustic infrastructure 
density. At $K=1$, lawnmower position RMSE grows  to approximately 350~m, rendering the 
precomputed lane structure ineffective and limiting 
area coverage to below 30\% regardless of fleet 
size. At $K=6$, structured lane coverage is 
restored, with coverage exceeding 90\%. The 
Voronoi-based strategy, by contrast, maintains 
RMSE below 20~m and stable area coverage across 
all tested beacon counts.

Further simulations quantify
this divergence across the full beacon range 
($K \in \{1,\dots,6\}$, $N=8$). A clear saturation 
threshold emerges at $K=3$ for both Voronoi-based 
and random walk strategies: beyond this point, 
additional beacons yield marginal improvements in 
AoI and RMSE. For lawnmower, no saturation is 
observed — coverage and AoI improve monotonically 
across the full tested range. It is also noted 
that random walk IVT remains low across all beacon 
counts despite area coverage not exceeding 65\%, 
confirming that IVT is an unreliable indicator of 
monitoring quality when interpreted in isolation 
from spatial coverage data.

\subsection{Comparative Strategy Analysis --- 
Global Profile}

\begin{figure*}[t]
\centering
\includegraphics[width=\textwidth]
{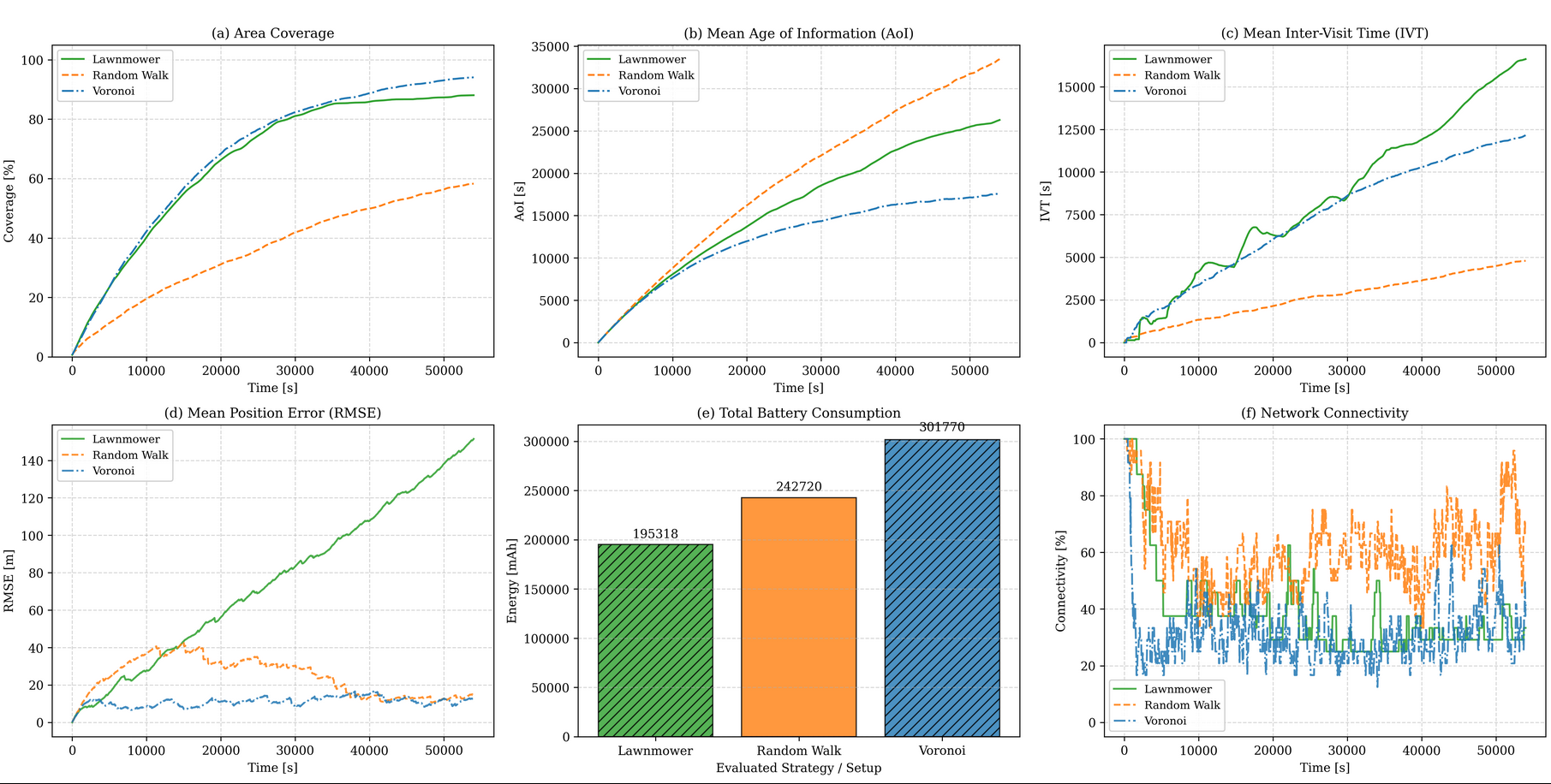}
\caption{Comparative performance of all three 
strategies ($N=8$, global profile, $K=4$): 
(a) area coverage, (b) mean AoI, (c) mean IVT, 
(d) position RMSE, (e) total battery consumption, 
(f) network connectivity. Voronoi-based achieves 
the lowest mean AoI ($\approx$700~s) with RMSE 
below 20~m. Lawnmower exhibits comparable area 
coverage but with unbounded AoI growth and 
eposition drift exceeding 140~m.}
\label{fig:strategy_comparison_global}
\end{figure*}

Fig.~\ref{fig:strategy_comparison_global} presents 
a direct performance comparison of all three 
strategies under a representative mid-range 
configuration ($N=8$, global profile, $K=4$). 
Voronoi-based achieves the lowest mean AoI of 
approximately 17500~s at $t = 52000$~s, compared 
to approximately 26000~s for lawnmower and 32000~s 
for random walk — a reduction of 33\,\% and 
45\,\% respectively. Concurrently, lawnmower 
position RMSE reaches 140~m, while both 
Voronoi-based and random walk maintain RMSE 
below 20~m after $t=38000$~s.

Despite achieving comparable area coverage to 
Voronoi-based ($\approx$87\,\% vs 90\,\%), 
lawnmower exhibits larger AoI growth, 
confirming that spatial coverage alone does not 
imply temporal coverage quality. The energy 
proxy $E_p$ reveals an expected inverse 
relationship between energy consumption and 
localization dependence: lawnmower consumes 
195,318~mAh against 301,770~mAh for 
Voronoi-based, reflecting the higher kinematic 
effort of adaptive replanning. This energy 
overhead is offset by the substantially superior 
temporal coverage, as captured by $\Psi$ in 
Section~\ref{sec:cost}.

\subsection{Comparative Strategy Analysis --- 
Turbine Inspection Profile}
\begin{figure*}[t]
\centering
\includegraphics[width=\textwidth]
{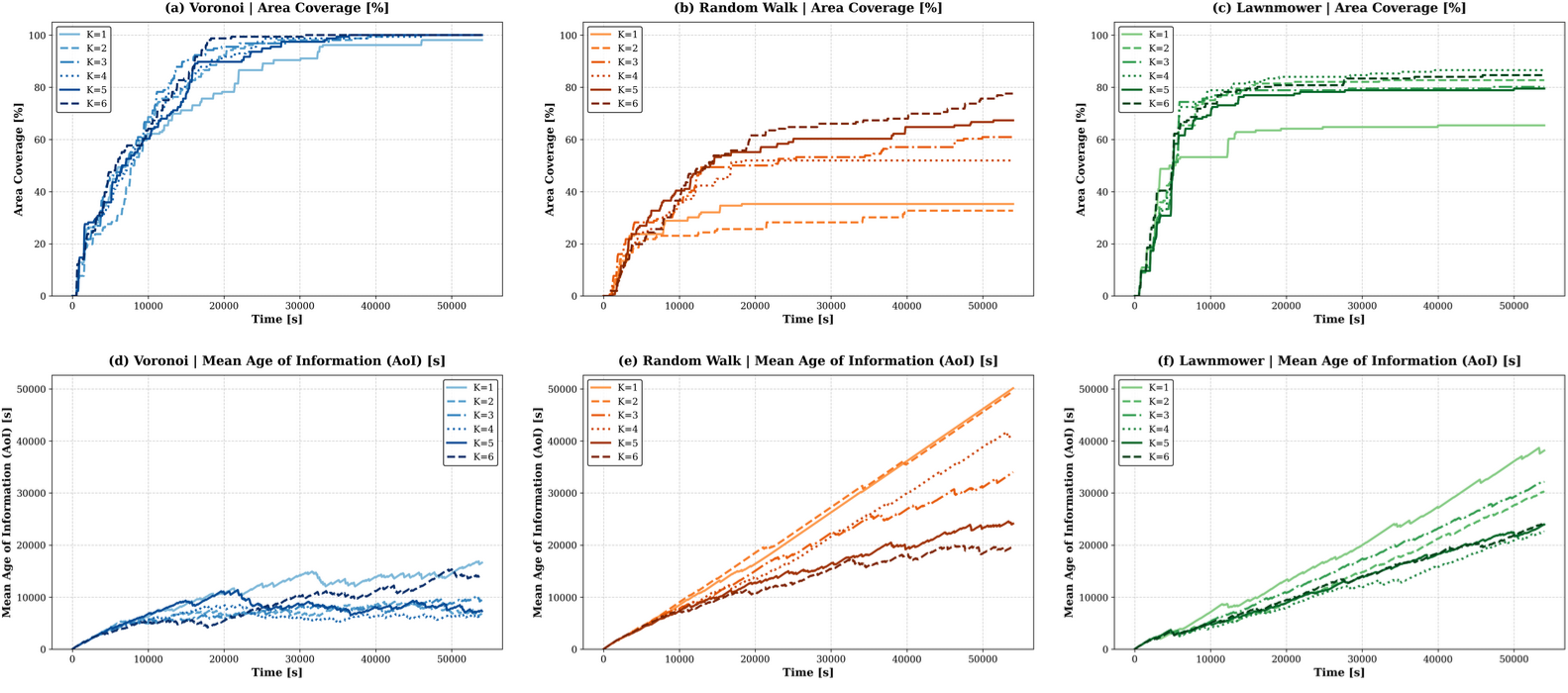}
\caption{Beacon scalability analysis across all three strategies ($N=8$, turbine inspection profile $K \in \{1,\dots,6\}$). 
}
\label{fig:beacon_scalability_turbines}
\end{figure*}
Fig.~\ref{fig:beacon_scalability_turbines} presents 
strategy performance under the turbine inspection 
profile ($N=8$, $K \in \{1,\dots,6\}$). The 
behavioral divergence between strategies is more 
pronounced than in the global profile. Voronoi-based 
achieves near-complete turbine coverage across all 
beacon counts, with a dramatic AoI reduction as 
beacon count increases — a effect driven by improved 
dead-reckoning correction in proximity to turbine 
structures used as sonar landmarks. Random walk 
maintains 85--90\,\% turbine coverage even at 
$K=1$, demonstrating robust fallback behavior 
under sparse infrastructure. Lawnmower coverage 
collapses below 20\,\% at $K=1$ and recovers 
only at $K \geq 4$, where it becomes competitive 
due to its ability to precisely follow perimeter 
inspection paths without redundant waypoints.

The energy distribution across turbine foundations 
further differentiates the strategies. Voronoi-based 
and random walk both exhibit uniform visitation 
across all 13 turbines, while lawnmower shows 
disproportionate energy concentration on a subset 
of foundations due to navigational drift — leaving 
some turbines entirely unvisited at low beacon 
counts. This confirms that for targeted inspection 
tasks under sparse infrastructure, adaptive 
strategies provide substantially more reliable 
coverage guarantees than deterministic approaches, 
regardless of fleet size.

\subsection{Fleet Size Scalability}

\begin{figure*}[t]
\centering
\includegraphics[width=\textwidth]
{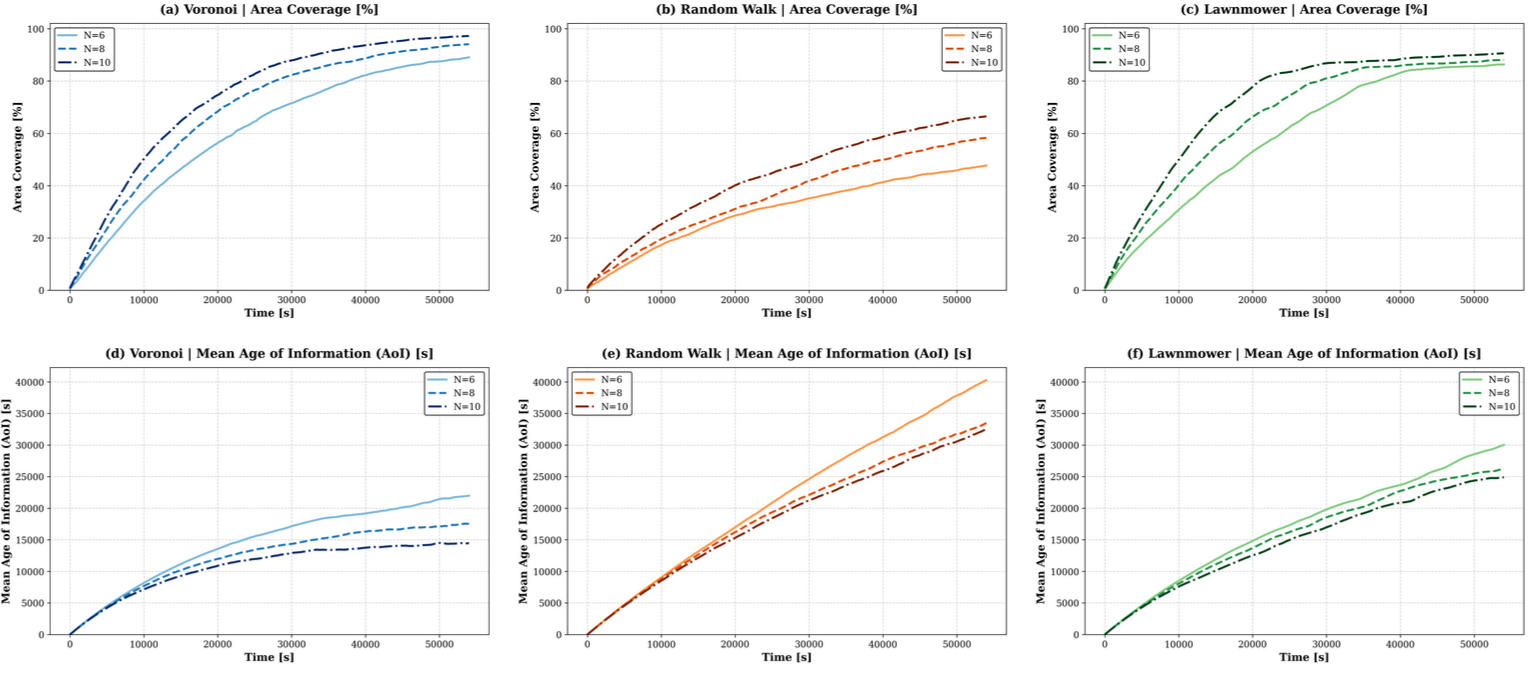}
\caption{Fleet's size scalability analysis across all 
three strategies ($N \in \{6,8,10\}$ AUVs, global profile, $K=4$)}
\label{fig:fleet_scalabillity}
\end{figure*}

Fleet scalability results for the global profile 
($K=4$, $N \in \{6, 8, 10\}$), as shown in Fig.~\ref{fig:fleet_scalabillity}, confirm that 
Voronoi-based coverage scales consistently with 
fleet size across all beacon counts, with each 
additional AUV providing meaningful AoI reduction. 
Diminishing returns become visible beyond $N=8$, 
where coverage curves begin to cluster, suggesting 
that workspace density rather than fleet size 
becomes the binding constraint at this scale.

Lawnmower fleet scaling is strongly mediated by 
beacon count. At $K=1$, adding AUVs provides 
negligible improvement — all fleet sizes achieve 
25--30\,\% coverage — confirming that navigational 
drift, not fleet density, is the binding constraint 
for deterministic strategies under sparse 
infrastructure. At $K \geq 4$, lawnmower scales 
effectively, achieving near-complete workspace 
coverage at $N=10$. Random walk coverage scales 
moderately with fleet size but remains below 
65\,\% across all tested configurations, 
reinforcing its unsuitability for global workspace 
coverage tasks regardless of fleet size or beacon 
count.

\subsection{Lifecycle Cost Analysis}

\begin{figure}[H]
\centering
\includegraphics[width=\columnwidth]
{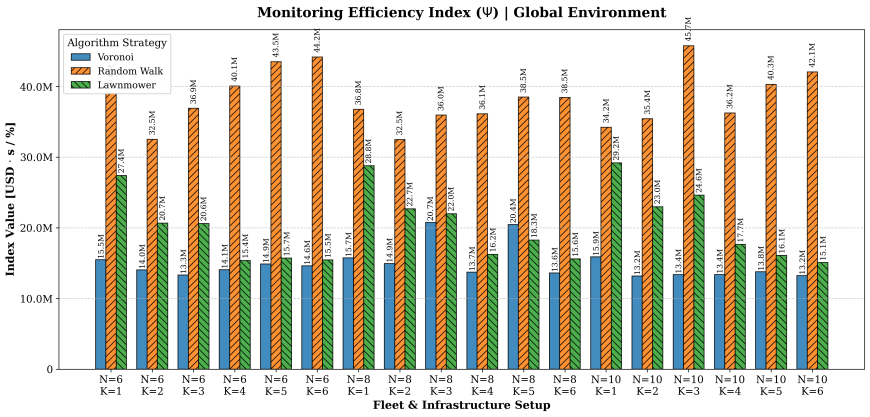}
\caption{Monitoring Efficiency Index $\Psi$ 
across all fleet and infrastructure 
configurations. Voronoi-based consistently 
achieves the lowest $\Psi$, remaining stable 
across beacon counts. A reduced Voronoi-based 
fleet ($N=6$) achieves 39--46\,\% lower 
lifecycle cost than a larger lawnmower fleet 
($N=10$) at $K \in \{2,3\}$.}
\label{fig:mei}
\end{figure}

Fig.~\ref{fig:mei} presents the Monitoring 
Efficiency Index $\Psi$ across all tested 
fleet and infrastructure configurations. 
Voronoi-based consistently achieves the lowest 
$\Psi$ across all configurations, with values 
remaining stable across beacon counts — 
confirming that its temporal coverage quality 
is largely independent of infrastructure 
investment beyond the $K=3$ saturation threshold. 
Random walk $\Psi$ increases substantially with 
beacon count, indicating poor return on 
infrastructure investment. Lawnmower $\Psi$ 
improves with beacon count but remains higher 
than Voronoi-based across all configurations.
\label{sec:cost}

To quantify the economic advantage of coordinated 
trajectory design, two representative scenarios 
are evaluated under sparse-to-moderate 
infrastructure:

\begin{itemize}
\item \textbf{Scenario 1} ($K=2$): A large 
lawnmower fleet ($N=10$) incurs a lifecycle 
cost of \$23.0M, compared to \$14.0M for a 
reduced Voronoi-based fleet ($N=6$) --- a 
reduction of 39.1\,\%.

\item \textbf{Scenario 2} ($K=3$): A large 
lawnmower fleet ($N=10$) incurs \$24.6M, 
compared to \$13.3M for a reduced Voronoi-based 
fleet ($N=6$) --- a reduction of 46.0\,\%.
\end{itemize}
These results directly validate the central 
hypothesis of this work: under sparse-to-moderate 
localization infrastructure, a coordinated 
partition-based strategy operating with a reduced 
fleet achieves superior temporal coverage at 
39--46\,\% lower lifecycle cost compared to a 
larger deterministic fleet.
\section{Discussion}

\subsubsection{Strategy Selection Guidelines}
The results establish clear operational guidelines 
for strategy selection based on mission profile 
and available infrastructure. For global workspace 
coverage under any beacon count, Voronoi-based 
partitioning provides the most reliable temporal 
coverage guarantees, with performance largely 
decoupled from localization accuracy beyond 
certain threshold. The biased random walk serves as a 
viable fallback under severely constrained 
infrastructure, particularly for turbine 
inspection tasks where its stochastic exploration 
achieves 85--90\,\% target coverage even at 
$K=1$. The lawnmower strategy is only recommended 
when two conditions are simultaneously satisfied: 
dense acoustic infrastructure and 
a requirement for near-complete workspace 
coverage or full turbine perimeter inspection, 
where its deterministic lane structure provides 
a systematic advantage.

\subsubsection{Localization Investment Guidelines}
A key practical finding of this work is that 
infrastructure investment beyond the certain
threshold yields diminishing returns for adaptive 
strategies. For operators deploying Voronoi-based 
or random walk fleets, increasing beacon count 
beyond a certain threshold provides marginal AoI improvement 
at substantial additional cost. Conversely, for 
lawnmower deployments, each additional beacon 
provides meaningful performance gains across 
the full tested range, making infrastructure 
investment more justifiable — but the absolute 
performance still remains inferior to adaptive 
strategies at equivalent cost. The MEI analysis 
confirms that reducing fleet size and investing 
in adaptive trajectory design yields better 
cost-performance trade-offs than expanding 
localization infrastructure for deterministic 
strategies.

\subsubsection{Limitations}
Several limitations of the current study should 
be acknowledged. First, the simulation environment 
is restricted to a 2D horizontal plane, omitting 
depth-dependent acoustic propagation effects and 
three-dimensional inspection requirements. Second, 
the fleet is assumed homogeneous, whereas 
real-world deployments may combine vehicles with 
different sensing payloads and kinematic 
constraints. Third, inter-vehicle communication 
is modeled as periodic and error-free within 
the defined range, abstracting away packet loss, 
latency, and acoustic channel contention that 
would degrade coordination in practice. Finally, 
the cost model uses fixed unit costs and a 
simplified energy proxy, which may not fully 
capture the operational economics of specific 
deployment scenarios.

\subsubsection{Future Work}
Future work will extend this framework to 
three-dimensional environments, incorporating 
depth-varying current models and acoustic 
shadow zones. Heterogeneous fleet compositions 
combining inspection-grade AUVs with faster 
relay vehicles for acoustic communication will 
be investigated. Additionally, the integration 
of real-time beacon placement adaptation — 
allowing the acoustic infrastructure to respond 
dynamically to observed navigation errors — 
represents a promising direction for further 
reducing the infrastructure requirements of 
deterministic strategies.
\section*{Acknowledgment}

The authors acknowledge the support of Marek Cygan, supervisor of the KN SPA student research group, and Elzbieta Jarzebowska, supervisor of the KN BLSPS student research group. The authors used ChatGPT (OpenAI) and Gemini (Google) for language editing and improving the clarity of the manuscript. This work was conducted at the Faculty of Physics and the Faculty of Mathematics, Informatics and Mechanics of the University of Warsaw.


\begin{thebibliography}{00}

\bibitem{rockmann2017operation}
C.~Röckmann, S.~Lagerveld, and J.~Stavenuiter, ``Operation and Maintenance Costs of Offshore Wind Farms and Potential Multi-use Platforms in the Dutch North Sea,'' in \emph{Aquaculture Perspective of Multi-Use Sites in the Open Ocean}, B.~H.~Buck and R.~Langan, Eds., 2017.

\bibitem{smith2011persistent}
S.~L. Smith, M.~Schwager, and D.~Rus, ``Persistent Robotic Tasks: Monitoring and Sweeping in Changing Environments,'' \emph{IEEE Transactions on Robotics}, 2011.

\bibitem{leonard2007collective}
N.~E. Leonard \emph{et al.}, ``Collective motion, sensor networks, and ocean sampling,'' \emph{Proceedings of the IEEE}, 2007.

\bibitem{alhabob2023age}
O.~A. Al-Habob \emph{et al.}, ``Age-Optimal Information Gathering in Linear Underwater Networks: A Deep Reinforcement Learning Approach,'' 2023.

\bibitem{cao2023joint}
X.~Cao \emph{et al.}, ``Joint optimization of AoI and energy for AUV-assisted data collection in underwater acoustic sensor networks,'' 2023.

\bibitem{hollinger2012multi}
G.~Hollinger and S.~Singh, ``Multi-robot coordination with periodic connectivity: Theory and experiments,'' \emph{IEEE Transactions on Robotics}, 2012.

\bibitem{zhou2023applications}
X.~Zhou \emph{et al.}, ``Applications of Voronoi Diagrams in Multi-Robot Coverage: A Review,'' 2023.

\bibitem{guo2023static}
J.~Guo \emph{et al.}, ``A Static Area Coverage Algorithm for Heterogeneous AUV Group Based on Biological Competition Mechanism,'' 2023.

\bibitem{mao2023cooperative}
S.~Mao \emph{et al.}, ``Cooperative Coverage Control for Heterogeneous AUVs Based on Control Barrier Functions and Consensus Theory,'' 2023.

\bibitem{paull2014auv}
L.~Paull \emph{et al.}, ``AUV Navigation and Localization: A Review,'' \emph{IEEE Journal of Oceanic Engineering}, 2014.

\bibitem{garau2005path}
B.~Garau \emph{et al.}, ``Path Planning for Autonomous Underwater Vehicles in Realistic Oceanic Current Fields: Application to Gliders in the Western Mediterranean Sea,'' 2005.

\bibitem{almuzaini2024trajectory}
K.~Almuzaini and A.~V. Savkin, ``Trajectory Planning for Autonomous Underwater Vehicles in Uneven Environments: A Survey of Coverage and Sensor Data Collection Methods,'' 2024.

\bibitem{prestero2001verification}
T.~Prestero, ``Verification of a six-degree of freedom simulation model for the REMUS autonomous underwater vehicle,'' Master's thesis, Massachusetts Institute of Technology and Woods Hole Oceanographic Institution, 2001.

\bibitem{freitag2005} 
L. Freitag, M. Grund, S. Singh, J. Partan, P. Koski, and K. Ball, ``The WHOI micro-modem: an acoustic communications and navigation system for multiple autonomous underwater vehicles,'' in \textit{Proc. OCEANS 2005 MTS/IEEE}, 2005,

\bibitem{ferreira2023sensor}
F. Ferreira, M. Pinto, and A. Matos, ``Sensor placement in an irregular 3D surface for improving localization accuracy using a multi-objective memetic algorithm,'' \textit{Sensors}, 2023.

\bibitem{kinsey2006survey}
J. C. Kinsey, R. M. Eustice, and L. L. Whitcomb, ``A survey of underwater vehicle navigation: Recent advances and new challenges,'' in \textit{IFAC Proceedings Volumes (7th IFAC Conference on Manoeuvring and Control of Marine Craft)}, 2006.

\bibitem{webster2012}
S.~E. Webster, L.~L. Whitcomb, and R.~M. Eustice, ``Advances in Single-Beacon One-Way-Travel-Time Acoustic Navigation for Underwater Vehicles,'' \emph{The International Journal of Robotics Research}, 2012.

\bibitem{wynn2014autonomous}
R.~B. Wynn \emph{et al.}, ``Autonomous Underwater Vehicles (AUVs): Their past, present and future contributions to the advancement of marine geoscience,'' \emph{Marine Geology}, 2014.

\end{thebibliography}
\end{document}